\documentclass{article}

\usepackage[preprint]{neurips_2026}

\usepackage[utf8]{inputenc}
\usepackage[T1]{fontenc}
\usepackage{hyperref}
\usepackage{url}
\usepackage{booktabs}
\usepackage{amsmath}
\usepackage{amsfonts}
\usepackage{nicefrac}
\usepackage{microtype}
\usepackage{graphicx}
\usepackage{xcolor}
\usepackage{tikz}
\usepackage{multirow}
\usetikzlibrary{shapes,arrows,positioning,fit,backgrounds}

\title{Procedural Knowledge Is Not Low-Rank:\\Why LoRA Fails to Internalize Multi-Step Procedures}

\author{%
  Simon Dennis \\
  i14, University of Melbourne \\
  \And
  Kevin Shabahang \\
  i14 \\
  \And
  Hao Guo \\
  i14 \\
  \And
  Rivaan Patil \\
  i14 \\
}

\begin{document}

\maketitle

\begin{abstract}
Parameter-efficient fine-tuning methods like LoRA have become the default for adapting large language models, succeeding across instruction following, style transfer, and factual adaptation. We show that for procedural knowledge---the ability to follow multi-step procedures with conditional branching through to terminal states---LoRA fails to match full fine-tuning at the ranks where it retains its efficiency advantage. In a systematic ablation ($r = 16$--$128$) on a procedural travel booking task (14 nodes), all LoRA configurations fail uniformly (task success $\leq 2.54$ vs.\ 4.11 for full fine-tuning, all $p < 0.001$), with scores \textit{decreasing} at higher ranks---despite maintaining 95--99\% conversation completion rates. Cross-domain replication on Zoom support (14 nodes) and insurance claims (55 nodes) at 8B confirms the failure generalizes: LoRA underperforms full fine-tuning by 0.8--2.2 points on average at both $r\!=\!32$ and $r\!=\!128$, with the largest gap on the most complex procedure. Quadrupling rank from 32 to 128 provides marginal improvement but does not close the gap. SVD analysis of the weight changes produced by full fine-tuning explains why: across three domains at both 3B and 8B, the mean effective rank of the update ranges from 761 to 1{,}026, and rank 128 captures only 43--51\% of the squared Frobenius norm. Together, these findings establish that for procedural tasks LoRA falls well short of full fine-tuning---a fundamental limitation for agentic applications.
\end{abstract}

\section{Introduction}

LoRA~\citep{hu2021lora} and its variants have become the dominant approach to fine-tuning large language models. By constraining weight updates to low-rank matrices, these methods reduce memory requirements by an order of magnitude while matching full fine-tuning on instruction following, style transfer, and factual adaptation tasks. The implicit assumption is that the knowledge acquired during fine-tuning lives in a low-dimensional subspace of the parameter space---that task-specific adaptations are low-rank perturbations to general language modeling.

We show that this assumption breaks down for procedural knowledge. When a model must learn to follow a multi-step procedure---navigating conditional branches, tracking implicit state across turns, and guiding conversations to terminal states---low-rank updates fail across the range of practical ranks ($r = 16$--$128$, up to 3.4\% of parameters). At sufficiently high rank LoRA trivially approaches full fine-tuning---LoRA at rank equal to the matrix's full rank is equivalent to full fine-tuning, and the LoRA parameter count exceeds full fine-tuning's once $r$ approaches $\min(m,n)/2$ for an $m \times n$ matrix---so the question is not whether \emph{some} LoRA rank suffices but whether LoRA succeeds at ranks where its parameter-efficiency advantage holds.

The distinction matters because procedural tasks are central to the emerging class of LLM-based agents. Customer support workflows, technical troubleshooting, booking systems, and diagnostic interviews all require models to follow structured procedures. Companion work~\citep{dennis2026compilation} establishes that full fine-tuning compiles such procedures into a small model's weights with near-frontier quality at two orders of magnitude lower cost than orchestration-based agents; whether parameter-efficient methods can do the same is this paper's question. If LoRA cannot capture procedural knowledge, then the efficiency gains that make LoRA attractive for deployment are unavailable for this class of applications.

Our central structural claim is that procedural knowledge is distributed across the full parameter space in a way that low-rank approximations cannot capture at practical ranks.

We present four lines of evidence:

\begin{enumerate}
    \item A \textbf{LoRA rank ablation} ($r = 16, 32, 64, 128$) on travel booking showing uniform failure across ranks ($p < 0.001$ vs.\ full FT), paradoxical degradation at higher ranks, and a clean dissociation between conversational competence and procedural competence (\S\ref{sec:lora}).
    \item \textbf{Cross-domain replication} of LoRA $r\!=\!32$ vs.\ full fine-tuning on Zoom support (8B) and insurance claims (8B, 55 nodes), confirming the failure generalizes across domains, model sizes, and procedure complexity (\S\ref{sec:lora}).
    \item \textbf{SVD analysis of weight diffs} across three procedural domains at both 3B and 8B, showing mean effective rank 761--1{,}026 and rank 128 capturing only 43--51\% of the squared Frobenius norm---direct evidence that the required parameter changes are high-rank (\S\ref{sec:svd}).
    \item \textbf{Training dynamics} across six epochs that rule out an optimization-failure explanation for the LoRA deficit: LoRA $r=32$ reaches \textit{lower} held-out per-token loss than full fine-tuning (0.81 vs.\ 0.89) while scoring worse on every behavioral criterion---it fits the surface token distribution more precisely without acquiring the state-to-action mapping the procedure requires (\S\ref{sec:dynamics}).
\end{enumerate}

\section{Method}

\subsection{Procedure Definition}

We represent procedures as directed graphs $F = (N, E, n_0, T)$ where $N$ is a set of nodes with role (agent/user) and prompt template, $E \subseteq N \times N \times C$ is a set of edges with optional conditions, $n_0 \in N$ is the start node, and $T \subseteq N$ is a set of terminal nodes (success, abandonment, escalation). This formalism captures the structure of procedural tasks: multi-turn conversations that follow a defined workflow from intake to resolution, with conditional branching at decision points.

We evaluate on three domains of varying complexity (flowcharts in Appendix~\ref{app:flowcharts}):

\textbf{Travel booking} (14 nodes, 3 decision hubs, 3 terminal states; Figure~\ref{fig:travel-flowchart}). The agent gathers travel preferences, presents destination options, negotiates selections, and confirms bookings. Decision hubs route based on information completeness, user satisfaction, and booking status. The flowchart includes cycles, yielding 86 unique acyclic paths of 4--17 turns.

\textbf{Zoom support} (14 nodes, 3 decision hubs, 3 terminal states; Figure~\ref{fig:zoom-flowchart}). A product-specific domain: the agent is a Zoom specialist who triages issues, walks users through diagnostic steps, and checks whether each step resolves the problem. The training data encodes domain-specific knowledge (Zoom's UI, settings, error codes) directly. The procedure yields 60 unique acyclic paths of 4--17 turns.

\textbf{Insurance claims} (55 nodes, 6 decision hubs, 4 terminal states; Figure~\ref{fig:insurance-flowchart}). The agent handles intake, claim type assessment, document gathering with completeness checking, coverage determination, settlement negotiation, and resolution. Nearly 4$\times$ the node count of the other domains, with nested loops and cross-phase dependencies. The procedure yields 2{,}381 unique acyclic paths of 9--39 turns.

\subsection{Data Generation}

We generate synthetic conversations by sampling paths through the flowchart and generating turn-by-turn dialogue with Claude Sonnet 4.5, varying scenario variables (destinations, budgets, user personalities, claim types) for diversity. This yields 1{,}912 training conversations for travel, 6{,}264 for Zoom (8 seeds $\times$ 783), and 2{,}700 for insurance. The model at inference sees only natural dialogue with no procedural annotations---the structure is implicit in how conversations flow.

\subsection{Fine-Tuning Configurations}

We compare two fundamentally different fine-tuning approaches:

\textbf{Full parameter fine-tuning.} We update all parameters of the base model with no parameter-efficient adapters. For the 3B model (Qwen 2.5 3B Instruct), training runs on a single RTX 5090 in bf16 precision with AdamW 8-bit, learning rate $2 \times 10^{-5}$ with cosine decay, and an effective batch size of 16 via gradient accumulation. We train for 20 epochs and select the best checkpoint by held-out eval loss; convergence is reached at epoch 4 and the score plateaus thereafter. For the 8B model (Qwen3-8B), training runs on $8 \times$ A100 with DeepSpeed ZeRO-3 (the full-precision optimizer state plus parameters does not fit on a single A100 80GB), learning rate $2 \times 10^{-5}$, and an effective batch size of 32. Zoom uses 10 epochs (best at epoch 2); insurance uses 20 epochs (best at epoch 3). Both 8B runs share the same base model and configuration.

\textbf{LoRA $r = 16, 32, 64, 128$.} We compare LoRA~\citep{hu2021lora} at four ranks against full fine-tuning. The adapters target all linear layers in the transformer block---attention q, k, v, and o projections plus the MLP gate, up, and down projections---with the base model held in bf16, scaling factor $\alpha = 2r$, and learning rate $2 \times 10^{-4}$ (the larger LoRA learning rate is standard practice and gives LoRA the best chance against full fine-tuning's $2 \times 10^{-5}$). We use a single LoRA learning rate; \S\ref{sec:dynamics} shows LoRA's held-out validation loss reaches its minimum at epoch 3 and rises thereafter---LoRA has converged (and begun overfitting), so a different learning rate or longer training would not rescue its behavioural performance. Adapters are merged into the base model before evaluation, so inference cost is identical to full-FT inference (no adapter overhead). The rank ablation ($r = 16$--$128$) runs on travel with the 3B model and the same training data used for the 3B full-FT comparison. The cross-domain replication uses LoRA at $r\!=\!32$ and $r\!=\!128$ on Zoom and insurance with the 8B model, matching the full fine-tuning base model and training data in each comparison.

\subsection{Evaluation Conditions}

For the LoRA rank ablation, we compare four LoRA ranks against full fine-tuning on travel booking (3B). For the cross-domain replication, we compare LoRA at $r\!=\!32$ and $r\!=\!128$ against full fine-tuning on Zoom and insurance (8B).

\subsection{Evaluation Protocol}

We use dynamic user simulation via Claude Sonnet 4.5, which generates contextually appropriate user responses based on conversation history and scenario variables. The user simulator has no knowledge of the underlying flowchart. We evaluate $n = 200$ scenarios per condition.

An LLM-as-judge methodology~\citep{zheng2023judging} scores each conversation on five criteria (1--5 scale):

\begin{itemize}
    \item \textbf{Task Success}: Did the agent execute the procedure correctly through to an appropriate terminal state, with consistent and accurate handling at each decision point?
    \item \textbf{Information Accuracy}: Did the agent correctly use and retain all user-provided information?
    \item \textbf{Consistency}: Did the agent maintain coherent state across the conversation?
    \item \textbf{Graceful Handling}: How well did the agent handle changes, ambiguity, and edge cases?
    \item \textbf{Naturalness}: Does the conversation read like talking to a skilled human agent?
\end{itemize}

We use Claude Sonnet 4.5 as the primary judge and verify all comparisons with an independent GPT-4.1 judge applying the identical rubric (\S\ref{sec:judge}, Appendix~\ref{app:gpt4-judge}).

Statistical comparisons use Wilcoxon signed-rank tests for paired conditions, Mann--Whitney $U$ tests for unpaired conditions, bootstrap 95\% confidence intervals (10{,}000 resamples, percentile method), Cohen's $d$ effect sizes, and Holm--Bonferroni correction across the five criteria ($\alpha = 0.05$).

\section{Results}

\subsection{Behavioral Failure Across Ranks}
\label{sec:lora}

\begin{table}[t]
\centering
\caption{LoRA rank ablation on travel booking ($n=200$ per condition, Claude Sonnet 4.5 judge). All LoRA configurations fail to internalize procedural knowledge despite maintaining high conversation completion rates ($p < 0.001$ vs.\ full FT on all criteria at all ranks). Scores \textit{decrease} at higher ranks, with the $r\!=\!128$ degradation significant ($p = 0.001$ vs.\ $r\!=\!64$). LoRA targets all linear layers, bf16 base model, $\text{lr} = 2 \times 10^{-4}$, $\alpha = 2r$, 20 epochs, merged into base before evaluation.}
\label{tab:lora}
\begin{tabular}{lcccccc}
\toprule
\textbf{Criterion} & \textbf{r=16} & \textbf{r=32} & \textbf{r=64} & \textbf{r=128} & \textbf{Full FT} \\
& {\scriptsize (13M, 0.4\%)} & {\scriptsize (26M, 0.9\%)} & {\scriptsize (52M, 1.7\%)} & {\scriptsize (104M, 3.4\%)} & {\scriptsize (3.1B, 100\%)} \\
\midrule
Task Success & 2.50 & 2.54 & 2.44 & 2.10 & \textbf{4.11} \\
Info.\ Accuracy & 2.21 & 2.27 & 2.08 & 2.10 & \textbf{4.75} \\
Consistency & 2.00 & 2.00 & 1.85 & 1.88 & \textbf{4.34} \\
Graceful Handling & 2.59 & 2.60 & 2.42 & 2.05 & \textbf{4.07} \\
Naturalness & 2.44 & 2.29 & 2.19 & 2.03 & \textbf{4.12} \\
\midrule
Completion rate & 95.5\% & 99.0\% & 96.5\% & 97.0\% & 100\% \\
\midrule
$d$ vs.\ Full FT & $-$2.38 & $-$2.33 & $-$2.53 & $-$2.90 & --- \\
\bottomrule
\end{tabular}
\end{table}

Table~\ref{tab:lora} presents the central result. The LoRA rank ablation reveals three properties of procedural knowledge that distinguish it from the tasks where LoRA typically succeeds.

\textbf{Uniform failure across ranks.} All four LoRA configurations score approximately half the full fine-tuning baseline: task success of 2.10--2.54 vs.\ 4.11, consistency of 1.85--2.00 vs.\ 4.34. Every LoRA-vs-Full-FT comparison is significant at $p < 0.001$ (Mann--Whitney $U$, Holm--Bonferroni corrected), with effect sizes $|d| > 1.9$ on all criteria at all ranks. Increasing the rank from 16 to 128---an 8$\times$ increase in trainable parameters, from 13M to 104M---produces no improvement. For comparison, LoRA at $r = 16$ matches full fine-tuning on instruction-following benchmarks~\citep{hu2021lora}; here, even $r = 128$ reaches only 51\% of full fine-tuning's task success score.

\textbf{Paradoxical degradation at higher ranks.} Scores \textit{decrease} monotonically beyond $r = 32$: task success drops from 2.54 ($r=32$) to 2.44 ($r=64$) to 2.10 ($r=128$). The $r\!=\!128$ degradation relative to $r\!=\!64$ is statistically significant on three criteria (task success $p = 0.002$, graceful handling $p < 0.001$, naturalness $p = 0.007$; Holm--Bonferroni corrected). The $r = 128$ condition, with 3.4\% of all parameters modified, achieves the \textit{lowest} task success of any rank. This counter-intuitive result suggests that higher-rank LoRA may overfit to surface patterns in the training data without capturing deeper procedural structure. If procedural knowledge were simply a high-rank version of the same kind of knowledge LoRA captures at low rank, we would expect monotonic improvement with rank. The degradation suggests a qualitative mismatch between LoRA's update geometry and the structure of procedural representations.

\textbf{Completion without correctness.} All LoRA ranks complete 95.5--99.0\% of conversations (i.e., reach some terminal state without breaking down), so the models do not collapse mid-conversation. But the low task-success scores (2.10--2.54 vs.\ 4.11) show that completion comes via the wrong paths---skipping decision hubs, missing required information, or reaching terminal states prematurely. LoRA preserves the base model's ability to sustain multi-turn dialogue but fails to add the procedural following that fine-tuning was meant to install. The dissociation is between \textit{completion} (the form of a conversation) and \textit{correctness} (following the procedure to the right terminal state via the right path).

\subsubsection{Cross-domain replication at 8B}

\begin{table}[t]
\centering
\caption{LoRA $r\!=\!32$ and $r\!=\!128$ vs.\ full fine-tuning across three domains ($n=200$ per condition, Claude Sonnet 4.5 judge). Each comparison uses the same base model and training data. Higher rank provides marginal improvement but does not close the gap to full fine-tuning. The deficit is largest on the most complex procedure (insurance, 55 nodes).}
\label{tab:lora-crossdomain}
\small
\begin{tabular}{llcccc}
\toprule
\textbf{Domain} & \textbf{Criterion} & \textbf{LoRA $r\!=\!32$} & \textbf{LoRA $r\!=\!128$} & \textbf{Full FT} \\
\midrule
\multirow{5}{*}{\shortstack[l]{Travel\\(3B, 1{,}912 train)}}
& Task Success       & 2.54 & 2.10 & \textbf{4.11} \\
& Info.\ Accuracy    & 2.27 & 2.10 & \textbf{4.75} \\
& Consistency        & 2.00 & 1.88 & \textbf{4.34} \\
& Graceful Handling  & 2.60 & 2.05 & \textbf{4.07} \\
& Naturalness        & 2.29 & 2.03 & \textbf{4.12} \\
\midrule
\multirow{5}{*}{\shortstack[l]{Zoom\\(8B, 6{,}264 train)}}
& Task Success       & 3.99 & 4.14 & \textbf{4.50} \\
& Info.\ Accuracy    & 3.38 & 3.33 & \textbf{4.26} \\
& Consistency        & 4.12 & 4.05 & \textbf{4.42} \\
& Graceful Handling  & 3.22 & 3.31 & \textbf{4.62} \\
& Naturalness        & 3.47 & 3.63 & \textbf{4.86} \\
\midrule
\multirow{5}{*}{\shortstack[l]{Insurance\\(8B, 2{,}700 train)}}
& Task Success       & 1.90 & 2.10 & \textbf{4.47} \\
& Info.\ Accuracy    & 2.73 & 2.79 & \textbf{4.40} \\
& Consistency        & 2.82 & 3.09 & \textbf{4.51} \\
& Graceful Handling  & 2.51 & 2.79 & \textbf{4.81} \\
& Naturalness        & 2.30 & 2.67 & \textbf{4.92} \\
\bottomrule
\end{tabular}
\end{table}

The rank ablation establishes the failure on travel booking; Table~\ref{tab:lora-crossdomain} confirms it generalizes across domains and model sizes. We train LoRA at $r\!=\!32$ and $r\!=\!128$ alongside full fine-tuning on Zoom support (8B, 6{,}264 conversations) and insurance claims (8B, 2{,}700 conversations) using the same base model (Qwen3-8B) and training data in each comparison.

Full fine-tuning dominates both LoRA ranks on every criterion in every domain. Increasing rank from 32 to 128---a 4$\times$ increase in trainable parameters---provides marginal improvement on Zoom ($+$0.06 average) and insurance ($+$0.24 average) but comes nowhere close to full fine-tuning. In insurance, LoRA $r\!=\!128$ achieves only 2.10 on task success vs.\ 4.47 for full fine-tuning---still near-total failure on the most complex procedure.

The pattern across domains is consistent: the gap to full fine-tuning is largest on insurance (avg gap 2.17 at $r\!=\!32$, 1.93 at $r\!=\!128$) and smallest on Zoom (avg gap 0.90 at $r\!=\!32$, 0.84 at $r\!=\!128$). On Zoom the gap is concentrated in graceful handling and naturalness; task success and consistency show smaller deficits, suggesting LoRA captures coarse procedural structure on this domain but fails on the nuanced, state-dependent behaviors that distinguish competent from excellent procedure following.

\subsection{Judge Robustness}
\label{sec:judge}

To test whether the behavioural results depend on the choice of judge, we re-score every condition with an independent GPT-4.1 judge using the identical rubric. Under the alternative judge, the direction of every comparison is preserved: full fine-tuning outperforms LoRA in all three domains and at every rank, the paradoxical degradation at higher LoRA ranks on travel replicates ($3.69 \to 3.64 \to 3.48 \to 3.10$ on Task Success), and the cross-domain ordering (insurance gap $>$ travel gap $>$ zoom gap) is unchanged. Magnitudes are smaller under GPT-4.1, particularly on Task Success, where Claude penalises substantively contradictory content while GPT-4.1 reads the criterion more literally as procedural-form completion; the two judges agree more closely on Information Accuracy and Consistency. Full per-condition numbers are in Appendix~\ref{app:gpt4-judge}.

\subsection{The Weight Update Is Genuinely High-Rank}
\label{sec:svd}

\begin{table}[t]
\centering
\caption{SVD analysis of weight diffs ($W_{\text{ft}} - W_{\text{base}}$) across 252 weight matrices per domain (36 of each layer type, one per transformer block). \textbf{Max} is the dimensional ceiling on rank, set by matrix shape; K/V projections are smaller because GQA reduces them (max rank 256 in 3B with 2 KV heads, 1024 in 8B with 8 KV heads). \textbf{$r$=128/Eff.} expresses LoRA's rank-128 budget as a fraction of the effective rank---a direct measure of LoRA-128's dimensional adequacy for the update. ``Energy at $r$'' = fraction of squared Frobenius norm captured by the top $r$ singular vectors of the full-FT update; by Eckart--Young, this upper-bounds what any rank-$r$ adapter could express.}
\label{tab:svd}
\small
\begin{tabular}{lccccccc}
\toprule
\textbf{Layer type} & \textbf{Max} & \textbf{Eff.\ rank} & \textbf{$r$=128/Eff.} & \textbf{Energy $r\!=\!16$} & \textbf{Energy $r\!=\!32$} & \textbf{Energy $r\!=\!128$} \\
\midrule
\multicolumn{7}{l}{\textit{Travel Booking (3B, 1{,}912 conversations)}} \\
MLP gate\_proj & 2048 & 1341 & 10\% & 13.9\% & 16.8\% & 26.7\% \\
MLP up\_proj & 2048 & 1303 & 10\% & 15.0\% & 17.9\% & 27.5\% \\
MLP down\_proj & 2048 & 1311 & 10\% & 16.3\% & 19.9\% & 30.4\% \\
Attn q\_proj & 2048 & 940 & 14\% & 12.4\% & 16.8\% & 33.0\% \\
Attn o\_proj & 2048 & 933 & 14\% & 12.7\% & 16.6\% & 32.4\% \\
Attn k\_proj & 256 & 193 & 66\% & 23.2\% & 33.9\% & 74.6\% \\
Attn v\_proj & 256 & 194 & 66\% & 22.6\% & 32.7\% & 72.9\% \\
\textbf{Overall} & --- & \textbf{888} & \textbf{14\%} & \textbf{16.6\%} & \textbf{22.1\%} & \textbf{42.5\%} \\
\midrule
\multicolumn{7}{l}{\textit{Zoom Support (8B, 6{,}264 conversations)}} \\
MLP gate\_proj & 4096 & 1009 & 13\% & 19.0\% & 25.6\% & 43.6\% \\
MLP up\_proj & 4096 & 981 & 13\% & 18.9\% & 25.5\% & 44.1\% \\
MLP down\_proj & 4096 & 1354 & 9\% & 14.3\% & 19.4\% & 36.1\% \\
Attn q\_proj & 4096 & 585 & 22\% & 21.6\% & 30.8\% & 55.3\% \\
Attn o\_proj & 4096 & 678 & 19\% & 21.2\% & 29.1\% & 51.2\% \\
Attn k\_proj & 1024 & 343 & 37\% & 25.7\% & 36.9\% & 64.6\% \\
Attn v\_proj & 1024 & 375 & 34\% & 25.3\% & 34.6\% & 60.4\% \\
\textbf{Overall} & --- & \textbf{761} & \textbf{17\%} & \textbf{20.9\%} & \textbf{28.8\%} & \textbf{50.8\%} \\
\midrule
\multicolumn{7}{l}{\textit{Insurance Claims (8B, 2{,}700 conversations)}} \\
MLP gate\_proj & 4096 & 1342 & 10\% & 13.7\% & 19.2\% & 36.2\% \\
MLP up\_proj & 4096 & 1367 & 9\% & 12.3\% & 17.7\% & 35.2\% \\
MLP down\_proj & 4096 & 1872 & 7\% & 7.4\% & 11.4\% & 26.4\% \\
Attn q\_proj & 4096 & 728 & 18\% & 17.3\% & 25.5\% & 49.5\% \\
Attn o\_proj & 4096 & 962 & 13\% & 12.9\% & 19.7\% & 41.9\% \\
Attn k\_proj & 1024 & 412 & 31\% & 20.7\% & 31.0\% & 59.3\% \\
Attn v\_proj & 1024 & 501 & 26\% & 16.6\% & 25.3\% & 52.5\% \\
\textbf{Overall} & --- & \textbf{1{,}026} & \textbf{12\%} & \textbf{14.4\%} & \textbf{21.4\%} & \textbf{43.0\%} \\
\bottomrule
\end{tabular}
\end{table}

The behavioral ablation shows that LoRA fails; SVD analysis of the weight changes explains \textit{why}. We compute $\Delta W = W_{\text{ft}} - W_{\text{base}}$ for every weight matrix in the model, perform singular value decomposition, and measure how much of the update energy (squared Frobenius norm) is captured at each rank (Table~\ref{tab:svd}).

\textbf{The SVD truncation is an upper bound on what any rank-$r$ adapter can capture.} By the Eckart--Young theorem~\citep{eckart1936approximation}, the optimal rank-$r$ approximation of any matrix is given by truncating its SVD to the top $r$ singular vectors; no rank-$r$ approximation can capture more squared Frobenius norm than this truncation. The ``Energy at $r$'' columns in Table~\ref{tab:svd} therefore upper-bound what \textit{any} rank-$r$ LoRA adapter could express of the full fine-tuning update, regardless of how it is trained. The actual capture by a trained LoRA is bounded above by these numbers (and is typically smaller, since LoRA optimizes a token-prediction loss rather than directly approximating the full-FT update). The interpretations below therefore describe the best case for LoRA, not its actual behaviour.

\textbf{The update is high-rank.} Across all 252 weight matrices in the travel 3B model, the mean effective rank is 888 and rank 128---the highest LoRA rank we tested---captures only 42.5\% of the update energy. This is direct evidence that the parameter changes required for procedural internalization are genuinely high-rank. Even an oracle rank-128 LoRA adapter would, at best, approximate less than half of what full fine-tuning achieves in each weight matrix.

\textbf{MLP layers are the LoRA-128 bottleneck.} In the travel model, the MLP projections (gate, up, down) have effective ranks of 1{,}300+, so rank-128 LoRA represents only $\sim$10\% of the rank needed to span the update; rank 128 captures only 27--30\% of the update energy. The MLPs implement the feed-forward computation that maps hidden states to output logits, and the high effective rank is consistent with procedural knowledge requiring distributed changes to how the model maps conversational state to next actions.

\textbf{Attention K/V are well-covered by rank 128 because the matrices are small.} The 3B model uses GQA with 2 KV heads, so K/V projection matrices have max rank 256 (vs.\ 2048 for MLP and Q/O), and their effective rank ($\sim$193) is correspondingly smaller. Rank 128 covers $\sim$66\% of K/V's needed rank and captures 73--75\% of K/V update energy---high in absolute terms, but a consequence of K/V matrices being smaller, not of procedural updates avoiding K/V. For LoRA at rank 128, the limiting factor is the MLP layers, where the required updates are dimensionally too far from rank 128 to fit.

\textbf{The pattern replicates across domains and model sizes.} We repeat the SVD analysis on the 8B Qwen3 models fine-tuned for Zoom support and insurance claims (Table~\ref{tab:svd}). The pattern is consistent: mean effective rank ranges from 761 (Zoom 8B) to 1{,}026 (insurance 8B), and rank 128 captures only 43--51\% of the update energy overall. Insurance shows the highest MLP effective ranks (down\_proj: 1{,}872, where rank 128 is only 7\% of needed), consistent with the 55-node procedure requiring the most distributed weight changes. Across all three domains, the MLP layers provide the LoRA-128 bottleneck.

\textbf{This rules out optimization-failure explanations.} Because the energy-at-$r$ numbers are an upper bound on what any rank-$r$ adapter could express, the LoRA deficit cannot be rescued by a different optimizer, learning rate, or initialization---even an oracle LoRA finding the SVD-optimal rank-128 subspace would still leave more than half of the full-FT update unrepresented in the MLP layers. The failure is structural: LoRA's rank budget is insufficient.

\subsection{Training Dynamics: LoRA Overfits Without Learning}
\label{sec:dynamics}

\begin{table}[t]
\centering
\caption{Per-turn cross-entropy loss across training epochs for full fine-tuning vs.\ LoRA $r=32$ ($n=50$ eval conversations). ``Early'' = first 4 turns, ``Late'' = last 4 turns. LoRA achieves \textit{lower} overall loss but worse behavioral outcomes (Table~\ref{tab:lora}). LoRA's late-turn loss rises after epoch 3 (overfitting), while full FT's decreases monotonically.}
\label{tab:dynamics}
\begin{tabular}{l|ccc|ccc}
\toprule
& \multicolumn{3}{c|}{\textbf{Full Fine-Tuning}} & \multicolumn{3}{c}{\textbf{LoRA $r=32$}} \\
\textbf{Epoch} & Early & Late & Avg & Early & Late & Avg \\
\midrule
0 (base) & 3.30 & 2.01 & 2.22 & 3.30 & 2.01 & 2.22 \\
1 & 0.33 & 1.11 & 0.99 & 0.30 & 1.02 & 0.91 \\
2 & 0.29 & 1.04 & 0.92 & 0.27 & 0.95 & 0.84 \\
3 & 0.28 & 1.02 & 0.90 & \textbf{0.25} & \textbf{0.92} & \textbf{0.81} \\
4 & 0.28 & 1.01 & 0.89 & 0.25 & 0.92 & 0.82 \\
5 & 0.28 & 1.01 & 0.89 & 0.25 & 0.94 & 0.84 \\
6 & \textbf{0.28} & \textbf{1.00} & \textbf{0.89} & 0.25 & 0.97 & 0.86 \\
\bottomrule
\end{tabular}
\end{table}

To understand \textit{when} style and structure are learned---and to test whether the LoRA deficit might be explained by insufficient optimization rather than a structural limit---we track held-out per-token cross-entropy across six epochs for full fine-tuning and LoRA $r=32$ ($n=50$ eval conversations), decomposing by conversation position (Table~\ref{tab:dynamics}). The loss numbers in this section are measured on held-out data, not the training set.

\textbf{LoRA achieves lower loss with worse behavior.} At convergence, LoRA's average held-out loss (0.81 at epoch 3) is lower than full fine-tuning's (0.89 at epoch 6). Yet the behavioral evaluation (Table~\ref{tab:lora}) shows LoRA at $r=32$ scoring 2.54 on task success vs.\ 4.11 for full fine-tuning. This rules out the most common counter-argument to the LoRA deficit---that LoRA simply underfit and could be rescued by more training or a different learning rate. An underfit model would have \textit{higher} loss and worse behavior. LoRA has \textit{lower} loss and worse behavior, which is the signature of capacity spent on the wrong objective: LoRA fits the surface token distribution more precisely by memorizing local patterns, without learning the state-dependent procedural structure that determines behavioral quality. Lower perplexity does not imply better procedure following.

\textbf{Early turns converge fast; late turns reveal LoRA's failure to track state.} Both methods learn early-turn predictions rapidly (3.30 $\to$ $\sim$0.28 by epoch 2) and early-turn loss is stable thereafter (0.25--0.28 at epochs 3--6 for both). Early turns---greetings, initial information gathering---are formulaic and predictable from local context, so they are stably modeled at low rank. Late turns require tracking accumulated state across the conversation, and this is where the two methods diverge: full fine-tuning's late-turn loss decreases monotonically (2.01 $\to$ 1.00), while LoRA's reverses after epoch 3 (0.92 $\to$ 0.97). The position-dependent overfitting---early-turn loss stable, late-turn loss rising---is the signature of LoRA fitting surface correlations that predict early tokens well but failing on the state-dependent predictions that later turns require.

\section{Discussion}

\textbf{Where does procedural knowledge live?} Four complementary results converge on an answer. The LoRA rank ablation (\S\ref{sec:lora}) shows that low-rank parameter updates cannot capture procedural knowledge on travel, and the cross-domain replication confirms this on Zoom and insurance at 8B---with the largest gap on the most complex procedure. The SVD analysis (\S\ref{sec:svd}) addresses \textit{representability}: the weight changes produced by full fine-tuning have mean effective rank 761--1{,}026, with MLP layers reaching up to 1{,}872---so even an oracle rank-128 approximation would capture less than half the required update. The training dynamics (\S\ref{sec:dynamics}) address \textit{learnability}: within LoRA's representable subspace, the optimizer converges to a point that fits held-out tokens \textit{better} than full fine-tuning while behaving worse, ruling out the possibility that LoRA merely underfit. These two arguments are independent---a parameterization could fail on representability but succeed on learnability, or vice versa---so showing both closes the loop on the LoRA deficit.

The SVD result is the most direct evidence. It shows that procedural knowledge is not merely ``hard to learn with LoRA''---it is distributed across hundreds of dimensions in every weight matrix, fundamentally exceeding LoRA's representational capacity. Rank 128 captures only 43--51\% of the update energy across three domains; rank 16 captures only 14--21\%. The weight changes that full fine-tuning makes are genuinely high-rank, and this property is consistent across domains of varying complexity (14--55 nodes) and model sizes (3B--8B).

The layer-type breakdown locates the LoRA-128 bottleneck. MLP layers---which compute the mapping from hidden state to output---have effective ranks up to 1{,}872; rank 128 covers only 7--13\% of the rank needed and captures only 26--44\% of the update energy. Attention K/V projections fit within rank 128 in absolute terms (capturing 53--75\% of energy), but largely because GQA constrains their effective rank to a few hundred dimensions; rank 128 covers 26--66\% of their needed rank. The substantive bottleneck is the MLP layers, where the required updates are dimensionally far above rank 128.

\textbf{A heterogeneous-rank LoRA might close the gap.} Standard LoRA practice applies the same rank to every targeted matrix, but the matrices have very different sizes and effective ranks. Our results suggest the LoRA budget should be allocated heterogeneously: assigning higher rank to MLPs (where rank 128 is far below what is needed) and lower rank to K/V (where rank 128 is already overkill) would cover MLPs much better at the same total parameter count. Adaptive-rank methods like AdaLoRA explore this direction; our results suggest the right intervention is specifically to favour MLP layers. We have not tested this.

This also explains the paradoxical degradation at higher LoRA ranks. Since even rank 128 captures less than half the required MLP updates, increasing rank provides diminishing returns on procedural knowledge while increasing capacity for surface-pattern overfitting.

\textbf{Limitations.} Three scope boundaries qualify the claims above.

\textit{Domain scope.} The three procedures studied---travel booking, Zoom support, and insurance claims---share a common structural pattern: intake, information gathering, decision routing, and resolution. Procedurally distinct task families such as clinical history-taking, legal discovery, or scientific protocol following may have different rank profiles. Our claim that procedural knowledge is high-rank is grounded in this customer-service pattern; whether it generalises to other procedural families is open.

\textit{Heterogeneous-rank LoRA is untested.} As argued above, the SVD analysis suggests that allocating LoRA budget heterogeneously---higher rank to MLP layers and lower rank to attention K/V---could close part of the gap at the same total parameter count. We do not test this. The ``LoRA fails'' result should therefore be read as ``uniform-rank LoRA at standard ranks fails,'' with a layer-aware variant remaining a plausible avenue for narrowing the gap.

\textit{Single base model family.} All experiments use Qwen 2.5 3B Instruct or Qwen3-8B. The high-rank profile of procedural updates in the SVD analysis could be specific to this family's architecture, training corpus, or post-training process. Replicating the analysis on other base models---e.g., Llama or Mistral---would establish whether the rank profile is universal or Qwen-specific.

\section{Related Work}

\textbf{Parameter-efficient fine-tuning.} LoRA~\citep{hu2021lora} and QLoRA~\citep{dettmers2023qlora} enable efficient adaptation by constraining updates to low-rank subspaces. These methods match full fine-tuning on instruction following, style transfer, and many NLP benchmarks. Agent-FLAN~\citep{chen2024agentflan} showed that naive agent fine-tuning entangles format following with reasoning. Our systematic rank ablation adds specificity: procedural behavior requires modifications too high-rank for parameter-efficient methods at the ranks where their efficiency advantage holds. The paradoxical degradation at higher ranks suggests that increasing rank within this regime can also increase capacity for surface-pattern overfitting.

\textbf{Procedural knowledge in LLMs.} \citet{hsiao2025procedural} formalized procedural knowledge as hierarchical task networks and showed it improves agentic workflows. SimpleTOD~\citep{hosseini2020simpletod} collapsed modular dialogue pipelines into single models. SynTOD~\citep{samarinas2024syntod} used state transition graphs for data generation. Companion work~\citep{dennis2026compilation} compiles flowchart-derived procedures into small fine-tuned models, achieving near-frontier quality at two orders of magnitude lower cost than orchestration; that work uses full fine-tuning, motivating the present paper's question of whether parameter-efficient methods can substitute. These papers establish that procedural knowledge benefits from fine-tuning but do not characterize \textit{how} it is represented or what distinguishes it from other fine-tuned knowledge.

\textbf{Knowledge distillation for agents.} FireAct~\citep{chen2023fireact} distilled GPT-4 ReAct trajectories into smaller models, achieving 77\% improvement. Agent Lumos~\citep{yin2024lumos} trained planning modules surpassing GPT-4. AgentTuning~\citep{zeng2024agenttuning} showed agent abilities can be internalized while maintaining general capabilities. \citet{hsieh2023distilling} demonstrated that a 770M model can outperform 540B PaLM through step-by-step distillation. These papers demonstrate effective distillation but do not examine whether LoRA can achieve the same transfer---our results suggest it cannot for procedural tasks.

\textbf{Task-oriented dialogue.} MultiWOZ~\citep{budzianowski2018multiwoz} established the benchmark paradigm for task-oriented dialogue. Our flowchart-based approach differs in generating training data from procedure specifications rather than collecting human dialogues, enabling systematic control over procedural coverage.

\section{Conclusion}

Procedural knowledge is not low-rank. A systematic LoRA ablation ($r = 16$--$128$) on travel and cross-domain replication at 8B on Zoom and insurance show LoRA underperforming full fine-tuning by 0.8--2.2 points on average despite sustaining multi-turn dialogue without breaking down (95--99\% completion). SVD analysis explains the mechanism: the full-fine-tuning update has mean effective rank 761--1{,}026 across three domains at both 3B and 8B, concentrated in MLP layers, so rank 128 captures only 43--51\% of the update energy. For practitioners, deploying agents that must reliably follow procedures requires full fine-tuning; the efficiency gains of parameter-efficient methods come at the cost of procedural competence.

\begin{ack}
\end{ack}

\bibliographystyle{plainnat}
\bibliography{references}

\appendix

\section{Procedure Flowcharts}
\label{app:flowcharts}

\begin{figure}[h]
  \centering
  \includegraphics[width=0.85\textwidth,keepaspectratio]{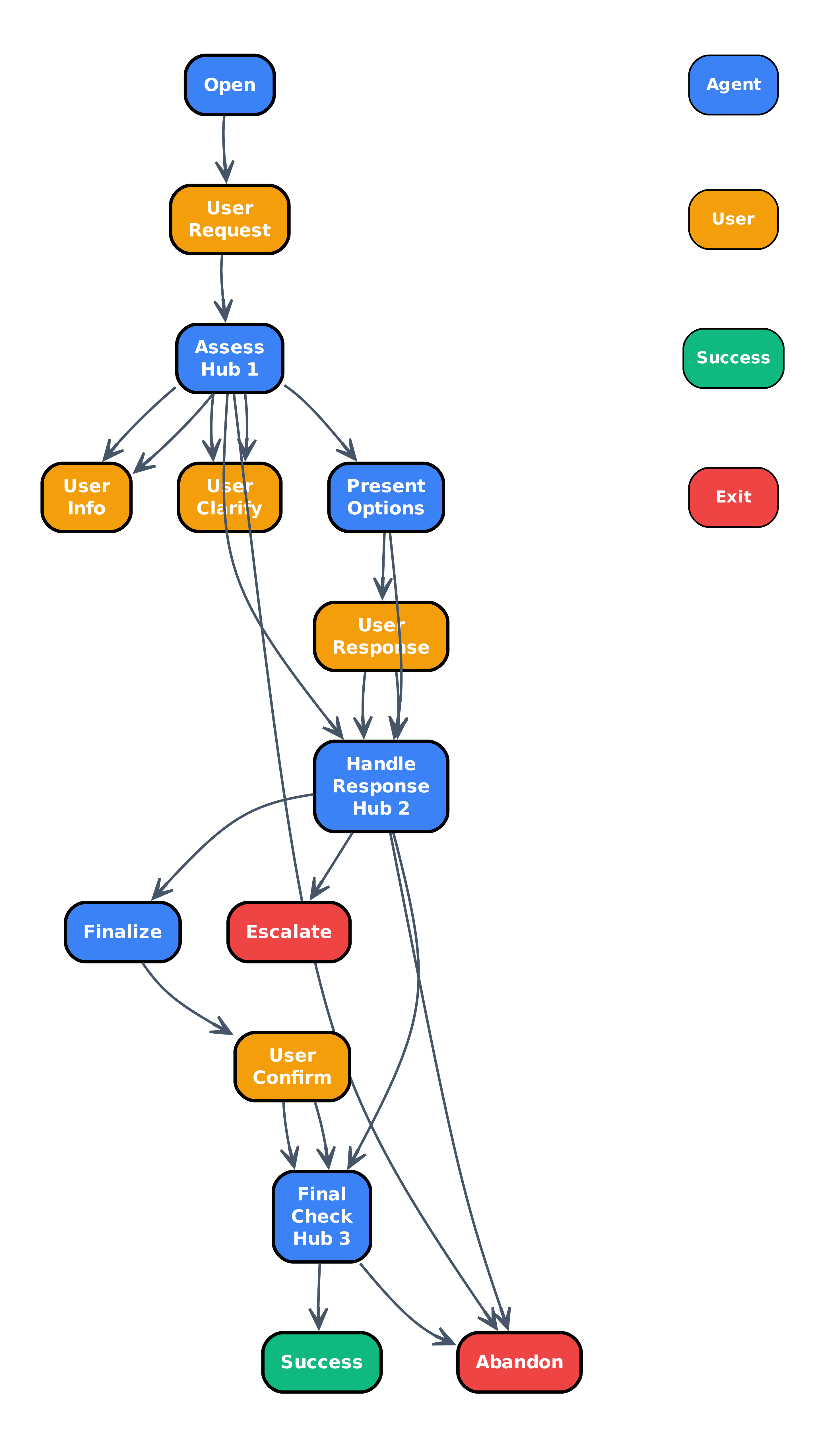}
  \caption{Travel booking procedure (14 nodes, 3 decision hubs, 3 terminal states). The agent opens, gathers requirements, routes through an assessment hub based on information completeness, presents options, negotiates, and either completes the booking or exits via abandonment.}
  \label{fig:travel-flowchart}
\end{figure}

\begin{figure}[h]
  \centering
  \includegraphics[width=0.85\textwidth,keepaspectratio]{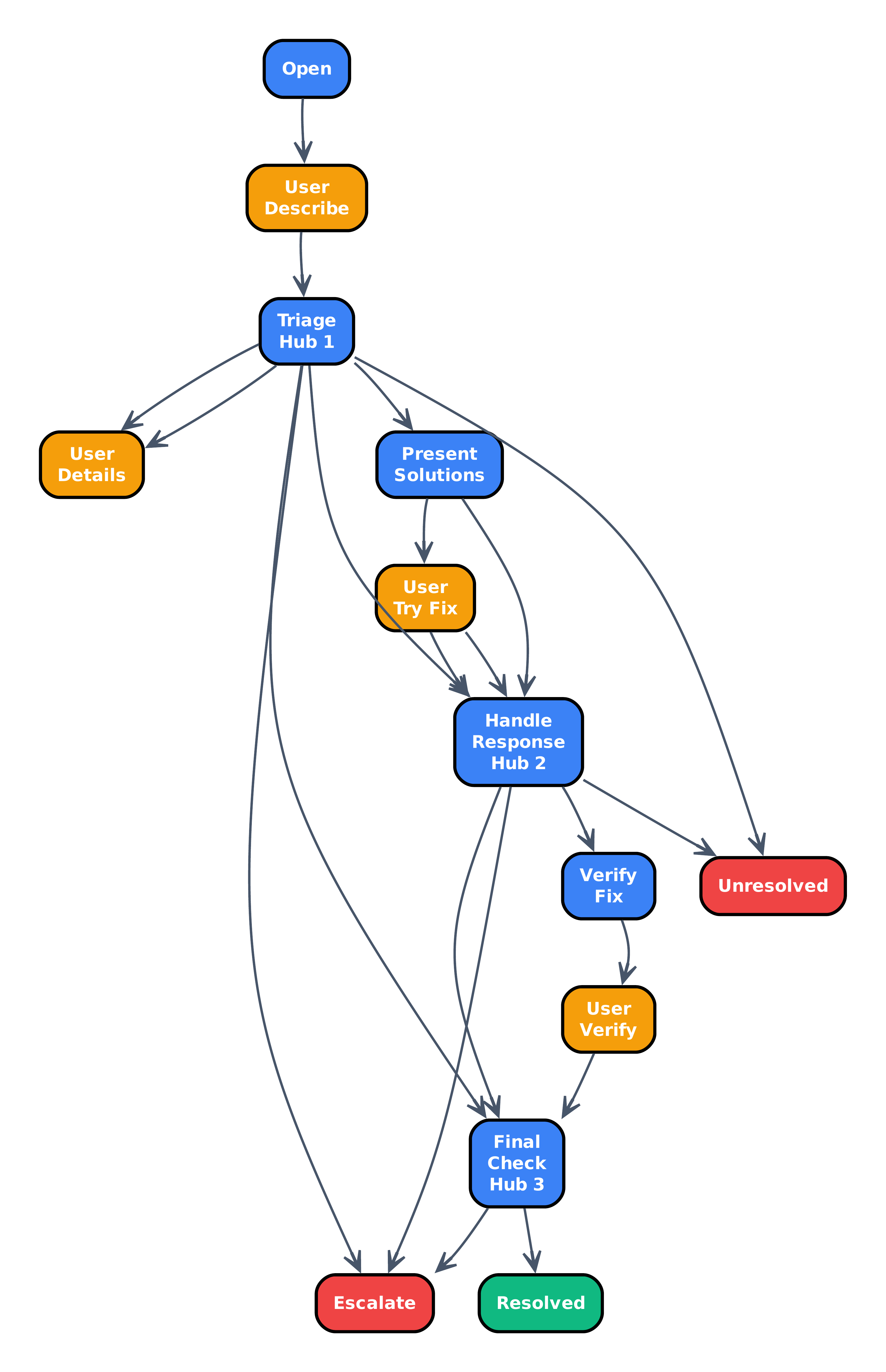}
  \caption{Zoom support procedure (14 nodes, 3 decision hubs, 3 terminal states). The agent triages the issue, walks the user through diagnostic steps, and verifies whether each step resolved the problem. The procedure terminates in \textsc{resolved}, \textsc{unresolved}, or \textsc{escalate}.}
  \label{fig:zoom-flowchart}
\end{figure}

\begin{figure}[p]
  \centering
  \includegraphics[width=\textwidth,height=0.9\textheight,keepaspectratio]{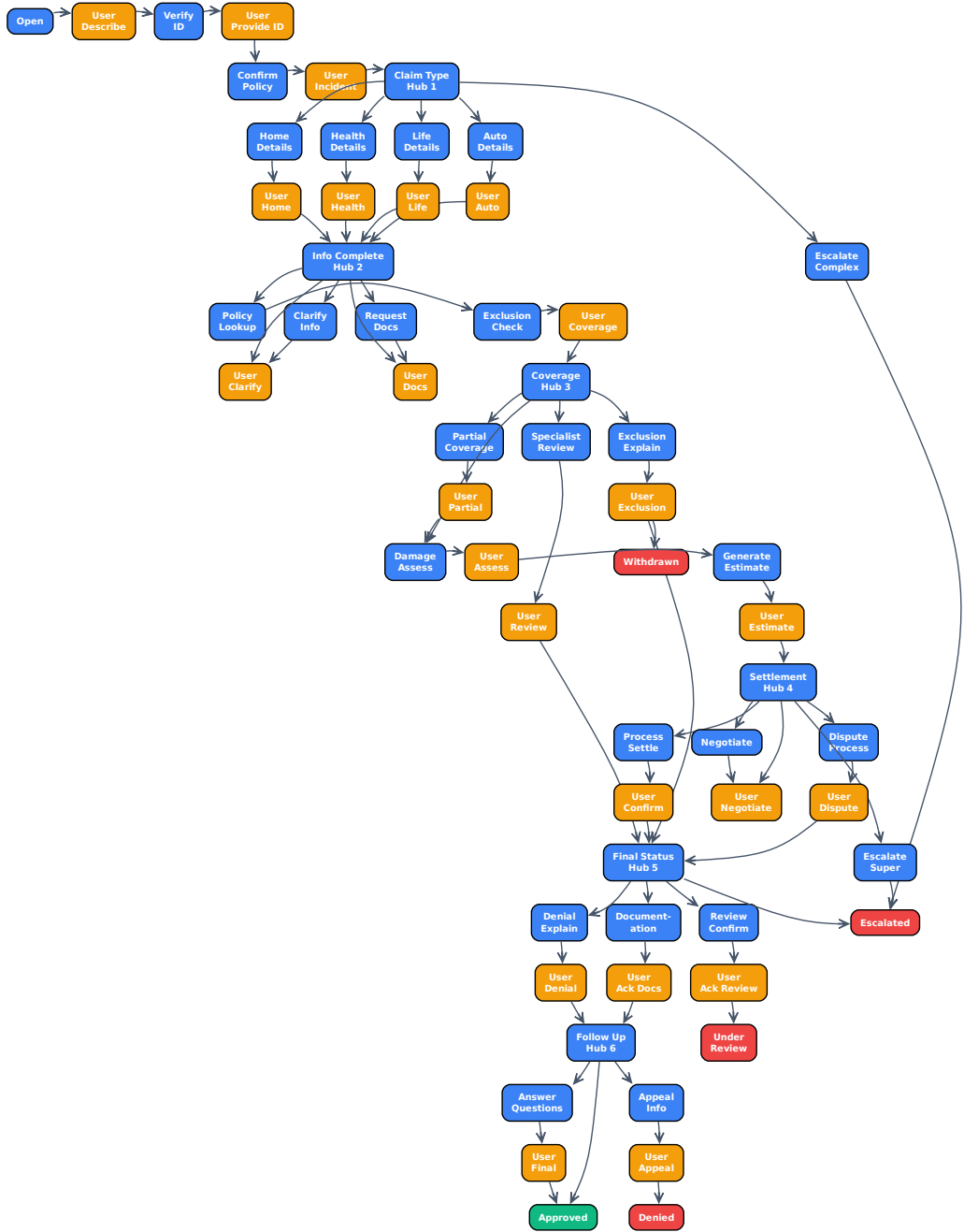}
  \caption{Insurance claims procedure (55 nodes, 6 decision hubs, 4 terminal states). Nearly $4\times$ the node count of the other two procedures, with nested loops across intake, document gathering, coverage determination, and settlement phases, plus cross-phase dependencies. This complexity yields 2{,}381 unique acyclic paths (9--39 turns) and is where the LoRA deficit is largest (Table~\ref{tab:lora-crossdomain}).}
  \label{fig:insurance-flowchart}
\end{figure}

Figures~\ref{fig:travel-flowchart}, \ref{fig:zoom-flowchart}, and~\ref{fig:insurance-flowchart} show the three procedures used in the experiments. Agent nodes are shown in blue, user nodes in amber, successful terminal states in green, and unsuccessful terminal states (abandonment, escalation) in red. Decision hubs are agent nodes with multiple outgoing edges; condition labels are suppressed for readability.

\clearpage

\section{Judge Robustness: GPT-4.1 Replication}
\label{app:gpt4-judge}

To test whether our findings depend on the choice of judge model, we re-scored every condition with an independent OpenAI GPT-4.1 judge using the identical rubric ($n=200$ per condition). Tables~\ref{tab:gpt4-rank-ablation} and~\ref{tab:gpt4-crossdomain} present the means.

\begin{table}[h]
\centering
\caption{GPT-4.1 judge replication---LoRA rank ablation on travel booking (3B). Cf.\ Table~\ref{tab:lora} (Claude Sonnet 4.5 judge). The qualitative pattern is preserved: full fine-tuning dominates every LoRA rank on every criterion, and increasing LoRA rank produces \emph{worse} scores (paradoxical degradation).}
\label{tab:gpt4-rank-ablation}
\small
\begin{tabular}{lccccc}
\toprule
\textbf{Criterion} & \textbf{Full FT} & \textbf{LoRA $r\!=\!16$} & \textbf{LoRA $r\!=\!32$} & \textbf{LoRA $r\!=\!64$} & \textbf{LoRA $r\!=\!128$} \\
\midrule
Task Success       & \textbf{4.32} & 3.69 & 3.64 & 3.48 & 3.10 \\
Info.\ Accuracy    & \textbf{4.11} & 3.37 & 3.41 & 3.19 & 3.18 \\
Consistency        & \textbf{4.29} & 3.33 & 3.37 & 3.13 & 2.91 \\
Graceful Handling  & \textbf{4.14} & 3.34 & 3.29 & 3.13 & 2.66 \\
Naturalness        & \textbf{3.91} & 2.92 & 2.85 & 2.73 & 2.46 \\
\bottomrule
\end{tabular}
\end{table}

\begin{table}[h]
\centering
\caption{GPT-4.1 judge replication---cross-domain comparison at 8B. Cf.\ Table~\ref{tab:lora-crossdomain} (Claude Sonnet 4.5 judge). Full fine-tuning leads on every criterion in every domain; the cross-domain ordering (insurance gap $>$ travel gap $>$ zoom gap) is preserved.}
\label{tab:gpt4-crossdomain}
\small
\begin{tabular}{llccc}
\toprule
\textbf{Domain} & \textbf{Criterion} & \textbf{LoRA $r\!=\!32$} & \textbf{LoRA $r\!=\!128$} & \textbf{Full FT} \\
\midrule
\multirow{5}{*}{Zoom (8B)}
& Task Success       & 4.34 & 4.43 & \textbf{4.56} \\
& Info.\ Accuracy    & 4.17 & 4.08 & \textbf{4.36} \\
& Consistency        & 4.54 & 4.52 & \textbf{4.82} \\
& Graceful Handling  & 3.05 & 3.10 & \textbf{3.23} \\
& Naturalness        & 3.77 & 3.83 & \textbf{4.00} \\
\midrule
\multirow{5}{*}{Insurance (8B)}
& Task Success       & 3.15 & 3.33 & \textbf{4.17} \\
& Info.\ Accuracy    & 3.71 & 3.81 & \textbf{4.38} \\
& Consistency        & 3.71 & 3.96 & \textbf{4.72} \\
& Graceful Handling  & 3.04 & 3.22 & \textbf{3.79} \\
& Naturalness        & 3.17 & 3.44 & \textbf{3.98} \\
\bottomrule
\end{tabular}
\end{table}

\textbf{Findings robust across both judges.} Three qualitative findings replicate cleanly: (1)~full fine-tuning outperforms LoRA on every criterion in every domain, at every rank, under both judges; (2)~the paradoxical degradation at higher LoRA ranks on travel is preserved (Task Success drops monotonically: $3.69 \to 3.64 \to 3.48 \to 3.10$ for $r=16,32,64,128$); (3)~the cross-domain pattern is preserved---the largest LoRA-vs-full-FT gap is on insurance (the most complex procedure), the smallest is on zoom.

\textbf{Magnitudes differ.} The average LoRA-vs-full-FT gap at $r\!=\!32$ across the five criteria is 1.7 (travel), 0.90 (zoom), and 2.17 (insurance) under Claude, versus 0.86, 0.22, and 0.85 under GPT-4.1---i.e., Claude shows gaps roughly $2\text{--}4\times$ larger. The gap is concentrated in the Task Success criterion, where Claude evaluates substantive procedural correctness (penalising contradictory or fabricated content even when stages are visited) while GPT-4.1 reads the criterion more literally as procedural form completion. On Information Accuracy and Consistency---criteria less open to interpretation---the two judges agree more closely.

\textbf{Implication.} The paper's claim that LoRA fails to internalise procedural knowledge holds under both judges: under either rubric reading, full fine-tuning produces measurably better procedural agents than LoRA at every rank tested. The Claude judge is stricter, particularly on substantive correctness within complex procedures, which amplifies the magnitudes reported in the main results. The qualitative conclusions and rank/complexity orderings are judge-independent.

\clearpage

\end{document}